\documentclass[runningheads]{llncs}
\usepackage[T1]{fontenc}

\usepackage{cite}
\usepackage{amsmath,amssymb,amsfonts}
\usepackage{graphicx}
\usepackage{textcomp}
\usepackage{xcolor}

\usepackage{t1enc}
\usepackage[utf8]{inputenc}
\usepackage{amsmath,amsfonts}
\usepackage{booktabs}
\usepackage{algorithm}
\usepackage{algpseudocode}
\usepackage{color}
\usepackage{multicol}
\usepackage{multirow}
\usepackage{url}

\begin{document}

\title{Describing Nonstationary Data Streams in~Frequency Domain}

\author{Joanna Komorniczak}

\authorrunning{}
\institute{\textit{Department of~Systems and~Computer Networks}\\
Wrocław University of~Science and~Technology, Wrocław, Poland\\
\email{joanna.komorniczak@pwr.edu.pl}}
\maketitle              
\begin{abstract}

\textit{Concept drift} is~among the~primary challenges faced by the data stream processing methods. The~drift detection strategies, designed to~counteract the~negative consequences of~such changes, often rely on~analyzing the~problem \emph{metafeatures}. This work presents the~\textit{Frequency Filtering Metadescriptor} -- a~tool for characterizing the~data stream that searches for the~informative frequency components visible in~the~sample's feature vector. The~frequencies are~filtered according to~their variance across all available data batches. The~presented solution is~capable of~generating a~\emph{metadescription} of~the~data stream, separating chunks into groups describing specific concepts on~its basis, and~visualizing the~frequencies in~the~original spatial domain. The~experimental analysis compared the~proposed solution with two \textit{state-of-the-art} strategies and~with the~\textsc{pca} baseline in~the~\textit{post-hoc} concept identification task. The~research is~followed by the~identification of~concepts in~the~real-world data streams. The~generalization in~the~frequency domain adapted in~the~proposed solution allows to~capture the~complex feature dependencies as a~reduced number of~frequency components, while maintaining the~semantic meaning of~data.

\end{abstract}

\section{Introduction}

The modern digital media generates exceptionally high volumes of~data that need to~be instantly processed by sophisticated solutions, often relying on~machine learning algorithms. \emph{Data stream processing} is~a~valid research area that considers the~data of~large volume and~high velocity, which makes solutions for data streams adequate for many modern problems~\cite{pigni2016digital}. An important factor in~processing data streams is~the~data nonstationarity, resulting from \emph{concept drifts}~\cite{agrahari2022concept}, which may lead to~the~loss of~the~method's recognition abilities.

The drift detection methods, designed to~recognize significant changes in~the data distribution, often rely on~the~\emph{metadescription} of~the~processed data. Methods may directly analyze the~classification quality of~the~classifier~\cite{gama2004learning}, or~other non-trivial factors, such as the~location of~class centroids~\cite{klikowski2022concept} and~a~set of~complex \emph{metafeatures}, precisely selected to~capture the~possible distribution variability~\cite{halstead2021fingerprinting}. While the~\textit{concept drift} became one of~the~primary difficulties faced by data stream processing methods, some other limitations related to~the~velocity and~volume of~the~data remain equally important. Those include the~processing of~data with \textit{high dimensionality}, often resulting in~the~reduced ability of~the~methods to~effectively classify data samples~\cite{hulten2001mining} and~recognize concept drifts~\cite{micevska2021sddm}. Another limitation, which has been recently frequently addressed, is~the~\textit{label delay}~\cite{grzenda2020delayed}. The~observation of~limitations related to~the~label access resulted in~the~proposition of~unsupervised drift detection methods~\cite{gemaque2020overview}.

Keeping in mind the actual applications of data streams, the~methods' evaluation on~real-wold data is~of~a~great significance~\cite{souza2020challenges}. This applies to~both \textit{static} data and~the~\textit{data stream} processing approaches~\cite{stapor2021design}. However, when considering the~task of~concept drift detection, the~moments of~concept changes are~often necessary to~compare the~evaluated approaches, since the~direct classification quality has been shown to~not reliably assess the~drift detection task~\cite{bifet2017classifier}. Therefore, providing the~\textit{explanation} of~concept drifts in~already collected data streams -- including their moments, severity and~dynamics~\cite{hinder2024one} -- could benefit the~quality of~method's evaluation. 

This work proposes the~\textit{Frequency Filtering Metadescriptor} -- a~method for unsupervised data stream characterization that extracts the~most informative frequency components visible in~the~feature vector of~each processed sample, approximated on~the~level of~a~data chunk. The~data analysis in~the~frequency domain allows for an effective generalization of~the~data with high dimensionality, while the~employment of~the~\emph{Fast Fourier Transform} -- the~effective extraction of~specific frequencies. The~proposed method is~described and~evaluated as a~\textit{post-hoc} processing tool. Such an approach allows its usage for the~purpose of~concept drift explanation~\cite{hinder2024one} or~annotation or~real-world data streams~\cite{krawczyk2017ensemble}. While the focus of the research is placed on \textit{post-hoc} analysis, the~presented method can be adapted to~incremental processing after the~preliminary analysis of~samples accumulated over an initial phase of~the~data inflow. This could be especially beneficial when processing data streams with recurring concepts \cite{gunasekara2024recurrent}, allowing for the~identification of~concepts occurring in~the~past with a~concise \textit{metadescription} of~data batches.

\paragraph{Contribution}

This work describes the~\textit{Frequency Filtering Metadescriptor} (\textsc{ffm}) -- a~method using frequency components of~high variance to~describe and~visualize the~nonstationary data streams. The~method analyzes the~data samples in~the~frequency domain, searching for informative components visible in~the~high-dimensional feature vector. The~particular benefit of~the~employed search strategy is~the~effectiveness and~generalization ability of~the~frequency domain in~the case of~high dimensional data.

The main contributions of~the~presented work are~as follows:
\begin{itemize}
    \item Proposition and~presentation of~the~\textsc{ffm} method for the~\textit{post-hoc} data stream characterization.
    \item The~data stream visualization approach based on~the~frequency components, allowing for the~visual assessment of~changes in~the~data.
    \item The~experimental analysis considering the~task of~unsupervised \emph{concept identification} with a~\emph{k-means} algorithm.
    \item Comparison with \emph{state-of-the-art} and~baseline \textit{metadescription} approaches employed in~drift detection methods and~classifier ensembles.
	\item The~presentation and~experimental evaluation of~the~strategy to~identify the~number of~concepts present in~the~data stream.
    \item The~\textit{concept identification} in~the~real-world data streams, including the~presentation of~data chunks in~the~\textit{metadescription} space and~the~concept membership.
\end{itemize}

\paragraph{Structure}

The rest of the work is organized as follows: Section \ref{sec-related} describes the~related works, focusing on the strategies of data stream \textit{metadescription} used in~the literature; Section \ref{sec-method} describes the method and expands on the intuition behind frequency analysis; Section \ref{sec-design} describes the design of experiments and their goals; Section \ref{sec-eval} analyses the obtained results; Section \ref{sec-insects} presents the analysis of real-world \textsc{insects} data streams \cite{souza2020challenges} with the proposed approach. Finally, Section \ref{sec-conclusions} concludes the work and shows possible future directions.

\section{Related works}\label{sec-related}

Processing data streams comes with inevitable challenges related to~the~volume of~the~data and~its \textit{temporal} nature~\cite{gama2010knowledge}. One of~the~most frequently addressed difficulties of~this data type is~the~data nonstationarity, resulting from \textit{concept drifts}~\cite{webb2016characterizing}. The significance of~recognizing concept changes stems from the~fact that they usually harm the~recognition quality of~methods since the~knowledge generalized in~machine learning models becomes outdated. 

The~primary axis of~the~concept drift taxonomy describes its impact on~the recognition model or, alternatively, the~data distribution shift in~relation to~the~decision boundary~\cite{gama2004learning}. Changes that do not affect the~recognition quality -- and~therefore cannot be recognized when monitoring the~quality of~the~model -- are~referred to~as \textit{virtual}. Meanwhile, those that affect the~decision boundary are~referred to~as \textit{real}~\cite{lobo2020spiking}. It is~worth keeping in~mind that the~potentially insignificant \textit{virtual} changes can be visible in~the~initial stage of~non-sudden \textit{real} concept changes~\cite{komorniczak2024structuring}. The~other axes of~the~concept drift taxonomy consider the~drift dynamics and~its recurrence. The~transition between the~consecutive concepts can be \textit{sudden} -- where one can see a~single time instant after which the~samples come from the~new concept. The~other categories describe slower-paced changes in~the~form of~\textit{gradual} or~\textit{incremental} drifts, in~which one can observe a~period of~concept transition. In~the~\textit{gradual} changes, the~samples in~the~transition period are~sampled from both the~previous and~the~emerging concepts, while in~the~\textit{incremental} changes, they form a~temporary superposition of~the~two transitioning concepts. Finally, regardless of~the~dynamics of~drift, the~concepts that appeared in~the~past may reoccur, which is~typical of~the~problems describing the~phenomena of~cyclic nature~\cite{gunasekara2024recurrent}.

\paragraph{Concept drift detection}

Since concept changes may have a~real effect on~recognition quality, it has become a~standard procedure to~monitor the~state of~a~system in~search for a~concept drift~\cite{sethi2015don}. For this purpose, many solutions have been proposed. The~initial drift detection methods exploited the~fact that concept drift affects the~classification quality. Those methods include the~\textit{Adaptive Windowing}~\cite{bifet2007learning}, which uses varying-width windows to~compare the~frequency of~errors. Methods that monitor the~quality of~a~classification model are~described as \textit{explicit}~\cite{gozuaccik2021concept}. The~primary benefit of~such a~drift detection approach is~the~possibility to~directly act upon a~change by adapting the~classification model to~the~current data. The~use of~\textit{explicit} methods also has some drawbacks. Since their operation is~based on~recognition quality, the~method will not be able to~detect the~\textit{virtual} drifts or~the~initial phases of~real ones that do not yet impact recognition quality. Another disadvantage is~the~reliance on~the~availability of~labels, which, in~the~data streams with high velocity, are~often delayed or~not available entirely~\cite{grzenda2020delayed}.

Another category of~methods monitor characteristics of~data stream processing other than those related to~the~quality of~the~model -- the~\textit{implicit} drift detection methods. Those include both supervised and~unsupervised approaches. The~supervised drift detectors can use labels to~monitor the~quality of~the~data distribution that are~not related to~the~errors made by the~classifier. The~\textit{Centroid Distance Drift Detector}~\cite{klikowski2022concept} is~a~simple yet effective approach that monitors the~class centroids to~detect concept changes. Labels are~also used in~the~\textit{Complexity-based Drift Detector}~\cite{komorniczak2023complexity}, which relies on~the~monitoring of~complexity measures~\cite{lorena2019complex} to~express the~difficulty of~the~classification task. Although the~supervised \textit{implicit} methods offer some independence from the~base classifier, they still rely on~access to~labels.

In the family of \textit{implicit} drift detectors, most of the methods are~\textit{unsupervised}. Those are~especially valuable in~the~context of~the~velocity of~the~data stream -- where the~time of~providing the~labels affects the~moment of~drift detection~\cite{komorniczak2024structuring}. Unsupervised methods can monitor quite a~wide range of~data characteristics, including the~data distribution analysis with hypothesis testing~\cite{sobolewski2013comparable}, or~the~percentage of~outliers measured with the~one-class classifier~\cite{gozuaccik2021concept}. Some interesting unsupervised methods utilize the~classification model but only to~measure label-independent characteristics of~the~underlying classification model. One of~the~most interesting ones of~this type is~the~\textit{Margin Density Drift Detector}~\cite{sethi2015don}, which examines the~distribution of~samples near the~decision boundary.  

All those characteristics considered in~drift detection -- from the~model quality and~its confidence to~the~temporal complexity of~the~classification task -- can be described as metafeatures of~the~data~\cite{komorniczak2024metafeatures}. This was directly addressed in~the~\textit{Meta-Feature-based Concept Evolution Detection} framework~\cite{guo2023meta}, where selected data distribution metrics captured the~statistical metafeatures of~the~data. Metafeatures were also used to~identify the~concept in~the~\textit{Fingerprinting with Combined Supervised and~Unsupervised Meta-Information} (\textsc{ficsum})~\cite{halstead2023combining}, where various metafeatures were used not only to~detect a~concept change but also to~re-identify it in~the~case of~recurrence. Some of~the~metafeatures used in~\textsc{ficsum} were previously used in~the~\textit{Feature Extraction for Explicit Concept Drift Detection}~\cite{cavalcante2016fedd}, which was dedicated to~time series analysis. The~measures included time series autocorrelation, partial autocorrelation, turning point rate, and~statistical measures: variance, skewness, and~kurtosis coefficient. 

An interesting strategy based on~analyzing the~frequency components of~data streams was used in~\textit{Multidimensional Fourier Transform}~\cite{da2017multidimensional}. The~authors extended the~\textit{unidimensional Fourier transform} to~detect changes in~frequencies and~amplitudes seen across many features over time. This strategy differs significantly from the~\textsc{ffm} since the~frequency components are~analyzed over time across specific features, which makes them suitable for time series analysis. In~contrast, the~proposed \textsc{ffm} approach searches for frequency components across the~feature vector characterizing each data sample. The~differences across those frequencies are~later used to~describe the~concepts visible in~the~data stream.

\paragraph{Data stream classification}

The drift detection task remains critical in~the~area of~data stream classification. The~proposed drift detection methods can serve as the~independent component of~a~processing pipeline or~be integrated with a~classifier, forming a~\emph{hybrid method}~\cite{wozniak2013hybrid}, which has become a~standard solution for data stream classification tasks.  Data stream classifiers often employ the~\emph{ensemble learning paradigm}~\cite{krawczyk2017ensemble}, profiting from the~possibility of~continuous modification of~the~ensemble's structure and~the~possibility of~integration with a~drift detection module. According to~the~taxonomy of~ensemble methods for data stream classification, the~\textit{active} ones use a~drift detection module and~directly act upon a~change. The~other category of~\textit{passive} methods incrementally adapts to~the~currently processed data, regardless if the~concept drift occurred or~the~data distribution remained stationary.

Among the~\textit{active} ensemble approaches, one should mention \textsc{adwin}\textit{Bagging}, which used an \textit{Adaptive Windowing} drift detector combined with \textit{online bagging} to~enable the~incremental learning of~classifier pool and~modification of~the~ensemble structure when the~concept drift is~detected~\cite{bifet2009improving}. Most of the methods utilized the~monitoring of~classification accuracy. Meanwhile, there exist ensemble approaches that, similarly to~\textit{implicit} drift detectors, base their detection on~other factors unrelated to~the~classification quality. One such method is~the~\textit{Covariance-signature Concept Selector}~\cite{ksieniewicz2023processing}, which examines the~covariance of~the~features to~detect concept changes and~to~select the~best model for current data distribution. A~similar strategy was used in~the~already mentioned \textsc{ficsum}~\cite{halstead2023combining}, which selects the~classifier dedicated to~the~currently solved task based on~the~gathered meta-information. Such a~selection is~especially valuable when processing the~data streams with recurring concepts, as it offers an opportunity to~use the~previous knowledge instead of~the~incremental adaptation from the~ground up.

\section{Method}\label{sec-method}

This work proposes a~\textit{Frequency Filtering Metadescriptor} (\textsc{ffm}) -- a~\textit{post-hoc} data stream processing method that describes the~data samples by filtering the~frequency components with the~largest variance. The~processing in~the~frequency domain allows for the~effective analysis of~data with high dimensionality -- since the~single frequency component can capture the~dynamics of~spatial features visible over the~entire original sample representation. This property of~frequency domain is~used in~the~data compression techniques, where the~low frequency components generalize the~complex spatial features~\cite{gonzales1987digital}.

The operation of~the~proposed approach is~described in~the~Algorithm~\ref{alg:pseudo}. A~single obligatory hyperparameter $n$ describes the~number of~selected informative frequency components. Another hyperparameter $c$~is~necessary only in~the~case of~concept identification by the~clustering algorithm and~describes the~number of~concepts present in~the~stream.

\begin{algorithm}[!htb]
\scriptsize
\caption{Pseudocode o the~\textit{Frequency Filtering Metadescriptor}}
\label{alg:pseudo}

\begin{algorithmic}[1]
\Statex $\mathcal{DS} = \{\mathcal{DS}_1, \mathcal{DS}_2, \ldots, \mathcal{DS}_k\}$ -- data stream

{\color{gray}\Comment{Hyperparameters}}

\Statex $n$ -- number of frequency components
\Statex $c$ -- number of concepts for clustering task

{\color{gray}\Comment{Parameters}}

\Statex $d$ -- dimensionality of samples
\Statex $R$ -- data stream \textit{metadescription}
\Statex $C$ -- concept identifiers
\Statex $I$ -- visualization of data stream


\vspace{1em}

\State $F_s \gets \emptyset$

    \ForAll{$\mathcal{DS}_k \in \mathcal{DS}$}
    
    \State $F_c \gets \emptyset$

        \ForAll{$X \in \mathcal{DS}_k$}
        {\color{gray}\Comment{Fourier transform on sample-level}}
        \State $X^{-1} \gets$ first $d/2$ values of $\mathcal{F}(X)$ real part
        \State $F_c \gets F_c \cup X^{-1}$
        \EndFor
    {\color{gray}\Comment{Frequency averaging}}
    \State $F_s \gets F_s \cup avg(F_c)$

    \EndFor
    {\color{gray}\Comment{Frequency selection based on variance}}

    \State $V \gets var(F_s)$
    \State $V_{max} \gets n $ frequencies of largest $V$
    
    \State $R \gets F_s[V_{max}]$
    {\color{gray}\Comment{Concept clustering}}

    \If {$C$ requested}
        \State $C \gets$ perform \emph{k-means} clustering of $R$ to $c$ clusters
        \State \textbf{return} $C$

    \EndIf
    {\color{gray}\Comment{Visualization}}

    \If {$I$ requested}
        \State $I \gets \emptyset$
        \ForAll{$R_k \in R$}

            \State $I_c \gets \emptyset$
            {\color{gray}\Comment{Filter selected frequencies}}
            \ForAll{$n_k \in 0,1,\ldots,n$}
            \State $I_n \gets R_k[V_{max}[n_k]]$ 
            {\color{gray}\Comment{Inverse transform into original domain}}
            \State $I_c \gets I_c \cup n$ first values of $ \mathcal{F}^{-1}(I_n)$
            \EndFor
        
        \State $I \gets I \cup I_c$
        \EndFor
        
        \State \textbf{return} $I$

    \EndIf
    
\end{algorithmic}
\end{algorithm}

The method produces the~\textit{metadescription} of~data $R$. Optionally, \textsc{ffm} can cluster the~data chunks into concept identifiers $C$ or~generate the~visual representation of~data chunks $I$ by presenting the~selected frequency components in~the~original input domain. 

At the~beginning of~data stream processing, the~frequency representation of~data stream $F_s$ is~empty (line 1). It is~iteratively extended by frequency representation of~data chunks $F_c$ (line 8). The~$F_c$ is~calculated as an average of~data samples in~the~frequency domain $X^{-1}$, obtained with Fourier transform $\mathcal{F}$ and~limited to~the~\textit{real} part of~the~complex result. Since the~result of~the~Fourier transform is~symmetric in~the~\textit{real} part, only the~first $d/2$ of~the~representation is~considered, with $d$ describing the~dimensionality of~a~sample. In~the~pseudocode, this process is~described in~lines 4:6. After the~generation of~$F_s$, the~variance of~specific frequencies is~calculated, and, based on~the~obtained result, the~$n$ frequency components with the~largest variance are~selected to~$V_{max}$ (lines 10:11). The~selected frequencies are~later used to~\textit{filter} the~complete set of~frequencies $F_s$ by limiting it to~$n$ components with the~largest variance. The~result is~stored in~variable $R$~as the~final \emph{metadescription} (line 12).

\begin{figure}[!b]
    \centering
    \includegraphics[width=0.7\linewidth]{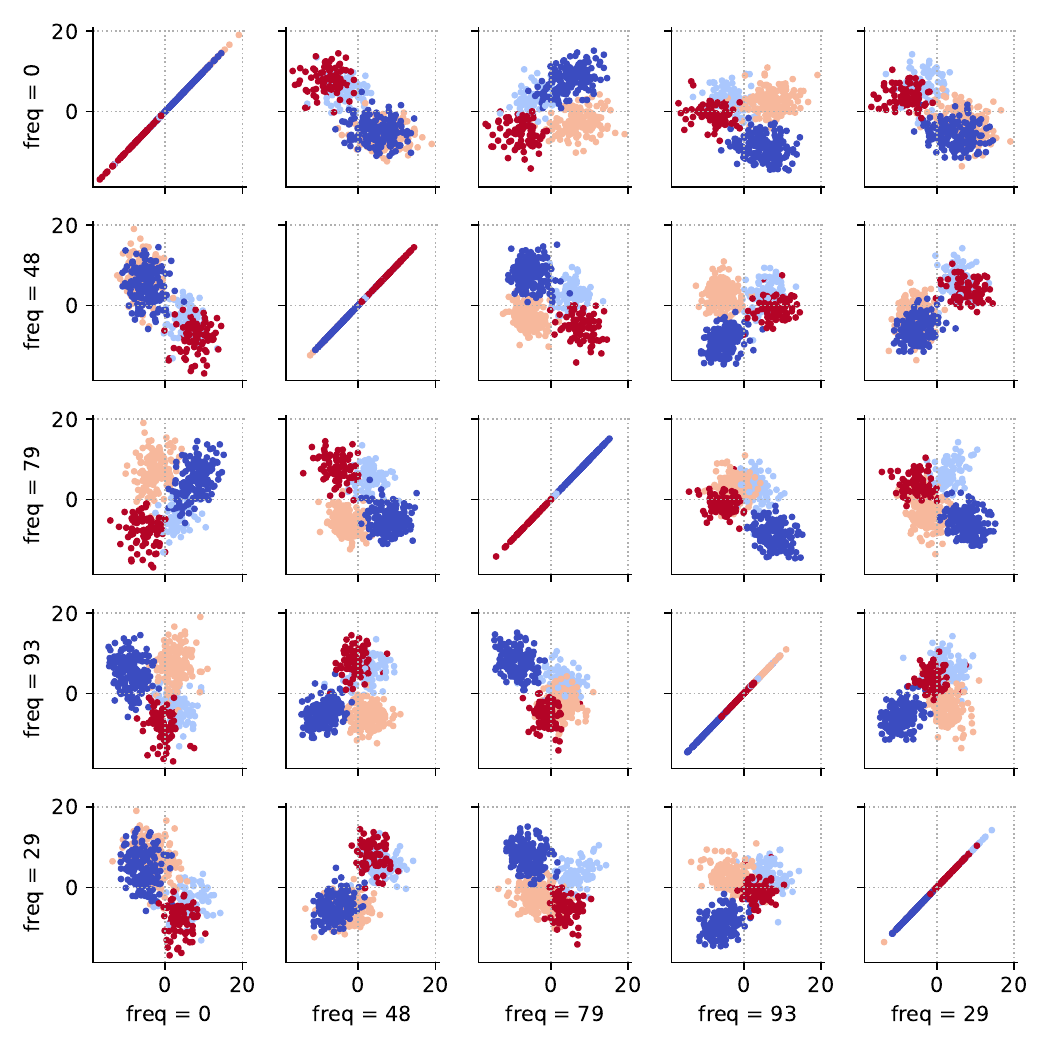}
    \caption{The data stream frequency representation clustered into four concepts based on~the~frequency representation. The~specific colors identify the~clusters obtained with \textit{k-means} algorithm.}
    \label{fig:cluster_example}
\end{figure}

If the~concept identifiers $C$ were requested, the~clustering of~normalized $R$~is~performed with a~\emph{k-means} algorithm. Here, the~second hyperparameter $c$~is~necessary for the~method to~divide the~chunk's \emph{metadescription} into groups representing specific concepts. The~clustering is~described in lines~13:16. The~result of~such concept identification based on~the~representation $R$~and $n=5$ is~presented in~Figure~\ref{fig:cluster_example}. 

The data stream used for this example was clustered into four concepts, described by various colors. Each point represents a~single data chunk. The~figure presents how the~chunks can be separated into clusters describing specific concepts using a~frequency representation of~chunks $R$. It is~important to~note that the~frequency of~$0$ will describe the~mean value of~the~feature vector. If such a~frequency is~selected, the~averaged value of~features was among the~$n$ selected as the~most informative metafeatures.

\begin{figure}[!htb]
    \centering
    \includegraphics[width=\linewidth]{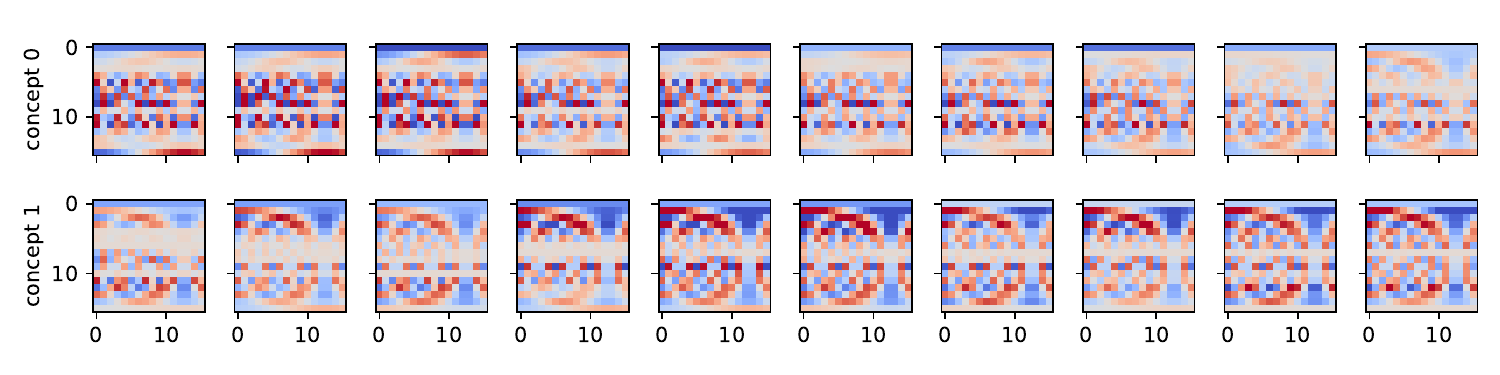}
    \caption{The visual representation of~data chunks, generated using the~$n=16$ frequency components. The~first row presents the~chunks from the~first part of~the~stream, and~the~second one -- the~chunks from the~following one. In~the~presented stream, the~gradual drift was injected, resulting in~a~smooth transition between concepts.}
    \label{fig:vis_example}
\end{figure}

If the~visual representation of~data chunks $I$ was requested, the~selected frequency components are~transformed with an inverse Fourier transform $\mathcal{F}^{-1}$ and~stacked in~rows to~form an image of~size $n$ x $n$. This process is~described by lines 19:28. Across all data chunks, the~specific frequencies from $V_{max}$ are~selected and~individually presented in~the~original spatial domain. The~result of~visualization of~a~data stream with a~single concept drift (i.e., two concepts) is~presented in~Figure~\ref{fig:vis_example}. 

The presented way of~processing offers the~possibility of~data generalization due to~the~extraction of~specific frequency components. The~discovery of~those components in~the~spatial domain would require the~inspection of~the~entire length of~the~sample's features. The~proposed \textsc{ffm} method significantly benefits from using the~\textit{Fast Fourier Transform} and~filtering in~the~frequency domain, allowing for the~computationally effective extraction of~frequency components. Furthermore, the~extracted \textit{metadescription} is~based solely on~the~data features, placing it in~the~\textit{unsupervised} category, making it resistant to~delayed or~limited labeling when employed in~real-world data stream setting.

It is~worth mentioning that the~presented processing scheme describes the~\textit{post-hoc} data stream analysis. The~entire data stream is~processed by extracting the~frequency components on~the~data instance level, averaging on~the~data batch level, and~ultimately, selecting the~final frequencies with the~largest variance on~the~data stream level. This type of~processing is~suitable for the~presented experimental analysis. However, it may not be adequate for the~\emph{incremental} data stream processing, where there is~no initial knowledge about the~processed data. It is~important to~note that it is~possible to~adapt the~processing scheme to~both \textit{batch} and~\textit{online} incremental processing of~the~data stream, including the~mechanism of~concept drift detection based on~extracted metafeatures. For the~presented research and~the~comparison with reference approaches, the~\textit{post-hoc} processing scheme will be adapted to~the~presented approach and~the~reference ones.

\section{Experiment design}\label{sec-design}

This section will describe the~setup and~the~goals of~the~experiments. The~presented approach aims to~enable concept identification in~nonstationary data streams. The~experiments use various types of~data streams, including data with extremely high dimensionality -- up to~500 features describing each sample. The~implementation of~the~method, the~experimental code, and~the~results are~publicly available as a~GitHub repository\footnote{\url{https://github.com/w4k2/FFM}}.

\subsection{Data streams}

The experiments were conducted using the~synthetic data streams generated using the~\textit{stream-learn} library~\cite{ksieniewicz2022stream}. 

The use of~synthetic data streams was motivated by the~possibility of~obtaining the~\emph{concept change ground truth}, indicating the~actual moments of~the~concept change and~the~concept identifier. Moreover, the~synthetic data stream generator enables the~specification of~a~wide range of~data stream characteristics, including the~data dimensionality and~the~number of~samples in~the~data stream. Finally, the~generation of~multiple data streams with the~same characteristics improves the~reliability of~the~results. The~detailed description of~generated data streams is~presented in~Table~\ref{tab:data-streams}. 

\begin{table}[h]
    \caption{Data stream generator configuration for the performed experiments}
    \centering
    \setlength{\tabcolsep}{5pt} 
    \renewcommand{\arraystretch}{1.1} 
    \scriptsize
    \begin{tabular}{l|l|l}
    \toprule
        \textbf{Experiment} & \textbf{Characteristics} & \textbf{Values} \\ \midrule
                 
        \multirow{3}{*}{Experiment 1} & number of~chunks & $500$ \\
        & chunk size & $50, 100, 200$ \\
        & number of~features &  $500$\\
        & number of~drifts & $1, 3, 5, 7, 9$ \\
        & drift type & $sudden$\\ \midrule
        
        \multirow{3}{*}{Experiment 2} & number of~chunks & $1000$ \\
        & chunk size & $256$ \\
        & number of~features &  $64$\\
        & number of~drifts & $3$ \\
        & drift type & $sudden, gradual, incremental$\\ \midrule
        
        \multirow{3}{*}{Experiment 3} & number of~chunks & $500$ \\
        & chunk size & $100,200,400$ \\
        & number of~features &  $500$ \\
        & number of~drifts & $1, 3, 5, 7, 9$ \\
        & drift type & $sudden$\\ \bottomrule
    
    \end{tabular}
    \label{tab:data-streams}
\end{table}

In the~first experiment, the~data streams were characterized by various chunk sizes -- from 50 to~200 samples in~each chunk -- and~various numbers of~drifts -- from a~single drift to~nine sudden concept changes throughout the~entire course of~the~stream. Regardless of~the~chunk size, the~stream consisted of~500~baches, and~the~samples were described by 500~features. In~the~second experiment, the~dimensionality of~data was limited to~64 features, which allowed for an experimental comparison with reference methods that were not well suited for high-dimensional data stream processing. The~data stream in~this experiment consisted of~1000 chunks with 256 samples each. Each stream had three concept drifts with various dynamics -- sudden, gradual, or~incremental. The~final experiment used data streams consisting of~500 chunks with various chunk sizes -- from 100 to~400 samples in~each data batch. The~dimensionality of~data was again set to~500 features. The~data streams used in~this experiment were characterized by various numbers of~sudden drifts -- from a~single change to~nine concept changes. Each stream type was replicated ten times to~enable statistical analysis of~the~results and~improve their stability.

\subsection{Goals of~experiments}

Three experiments were designed to~thoroughly evaluate the~\textsc{ffm} method in~various data stream environments and~compare the~presented approach with \textit{state-of-the-art} and~baseline solutions for \textit{metadescription} of~the~data stream. 

\paragraph{Selecting number of~frequency components}

The first experiment aimed to~evaluate the~influence of~an $n$ hyperparameter on~the~operation of~the~method. The~examined value describes the~number of~frequency components considered in~the~concept identification task. The~experiment evaluated five values, from analyzing a~single frequency component to~16 components with the~largest variance. Additionally, since selecting components is~based on~averaging the~samples across data chunks, the~experiment evaluated three data chunk sizes from 50 to~200. The~larger size of~the~data chunk should allow for a~better generalization of~frequency components and, hence, could allow for obtaining a~better representation of~a~data stream.

The representation of~the~data stream obtained with \textsc{ffm} was normalized and~clustered with \emph{k-means} to~an actual number of~concepts observed in~the stream. The~number of~concepts is~equivalent to~the~number of~drifts incremented by 1. After the~clustering, the~obtained concept identifiers were compared with \emph{concept ground truth}, identifying the~actual concepts present in~the~specific point of~the~data stream. The~\emph{normalized mutual information} clustering metric was used in~this experiment. 

Selecting more frequency components is~expected to~allow for a~more precise concept identification. However, the~experiment searches for a~minimal $n$ offering satisfactory results since the~data of~higher dimensionality poses particular challenges across many machine learning tasks~\cite{verleysen2005curse}.

\paragraph{Comparison with reference approaches}

The second experiment was designed to~compare the~\textit{metadescription} extracted with \textsc{ffm} with the~ones used by \textit{state-of-the-art} data stream classification and~drift detection strategies. The~following approaches were evaluated:
\begin{itemize}
    \item [\textsc{ced}] the~metafeatures used in~the~\emph{Meta-Feature-based Concept Evolution Detection framework}~\cite{guo2023meta}. Those included statistical measures such as mean, standard deviation, correlation, skewness, and~kurtosis. Those five metafeatures were calculated for each of~the~original attributes and~later aggregated using the~mean and~standard deviation as a~summarization function~\cite{rivolli2022meta}. This resulted in~a~total number of~10 metafeatures describing each data chunk.

    \item [\textsc{ici}] the~set of~metafeatures from different categories, selected based on~the~research in~\emph{On metafeatures ability of~implicit concept identification}~\cite{komorniczak2024metafeatures}. The metafeatures included \textsc{int} index~\cite{bezdek1998some}, normalized relative entropy, maximum Fisher's discriminant ratio~\cite{lorena2019complex}, class concentration coefficient, target attribute Shannon's entropy, joined entropy, mutual information, the~performance of~the~worst decision tree node, mean, median and~trimmed mean. Similar to~the~approach used when calculating metafeatures for \textsc{ced}, when possible, mean and~standard deviation were used as summarization functions. The~\textsc{ici} approach resulted in~a~total number of~19 metafeatures describing each data chunk.

    \item [\textsc{ffm}] the~method proposed in~this work, analyzing the~frequency components with the~largest variance. The~$n$ value in~this experiment was set to~8, resulting in~the~analysis of~8 metafeatures.

    \item [\textsc{pca}] the~baseline approach, extracting two principal components from original features.

\end{itemize}

Similarly to~the~approach adopted in~the~first experiment, the~obtained representation was normalized and~clustered with \emph{k-means} to~an actual number of~concepts present in~the~data stream, equivalent to~$4$ (number of~drifts $+ 1$). The~cluster identifiers from \emph{k-means} were then compared with actual concept identifiers. In~this experiment, the~number of~evaluation metrics was extended to~four: \textit{normalized mutual information} (\textsc{nmi}), \textit{adjusted Rand score}, \textit{completeness} and~\textit{homogeneity}~\cite{amigo2009comparison}.

The observations from this experiment should determine whether the~proposed \textsc{ffm} method is~competitive with \textit{state-of-the-art} solutions employing the data \textit{metadescription}. Since the~proposed approach is~the~only one that analyses samples in~the~frequency domain, it should primarily show its advantages when processing data with high dimensionality. Meanwhile, in~case of~a~significant number of~features, the~remaining methods will suffer from high computational and~memory complexity, forcing the~dimensionality limit of~the~processed data stream to~64 features.

\paragraph{Identifying the~number of~concepts}

The final experiment focused on~the~proposed \textsc{ffm} approach, evaluating its ability to~describe the~high-dimensional data streams by identifying the~number of~concepts present in~the~given period. The~ability of~methods not only to~detect concept changes but also to~identify the~number of~concepts occurring in~the~stream can be of~great significance when processing data streams with recurring concepts -- where after a~concept change, a~concept from the~past can reoccur~\cite{gunasekara2024recurrent}. Identifying the~number of~concepts and~the~moments of~their occurrence could become valuable in~designing ensemble methods for data stream processing, allowing for the~dynamic classifier fusion~\cite{zyblewski2020dynamic}.

In this experiment, the~number of~concept changes in~the~clustering process was unknown. The~proposed approach used the~clustering with \emph{k-means} to~a~range of~target cluster numbers $c$~and assessed the~obtained results with \emph{silhouette score}~\cite{shahapure2020cluster,rachwal2023determining}. Since this measure does not require actual cluster labels, the~experiment did not utilize the~\emph{concept ground truth} to~assess the~method. The~satisfactory results of~this experiment could motivate the~use of~\textsc{ffm} and~representation clustering to~identify the~number of~concepts in~the~data stream. 

In this experiment, the~streams were characterized with from 1 to~9 concept drifts -- i.e., from 2~to~10 concepts. The~search space for the~number of~concepts (used as a~number of~clusters in~\emph{k-means}) aggregated all possible values from 2 to~11. As an additional factor, the~various chunk sizes were evaluated. Similarly to~the~first experiment, the~larger chunk size should allow for a~more precise data stream description.

Additionally, in~this experiment, the~visualization of~data chunks is~presented, showing the~selected frequency components visible in~the~data. This presents how \textsc{ffm} could aid the~visual assessment of~the~processed nonstationary data. This experiment used the~hyperparameter value of~$n=16$.

\section{Experimental evaluation}\label{sec-eval}

This section analyzes the~results of~the~conducted experiments, shows the~limitations and~opportunities of~the~proposed \textsc{ffm} approach, and~compares it with reference methods of~data stream \textit{metadescription}.

\subsection{Selecting number of~frequency components}

\begin{figure}[!htb]
    \centering
    \includegraphics[width=0.9\linewidth]{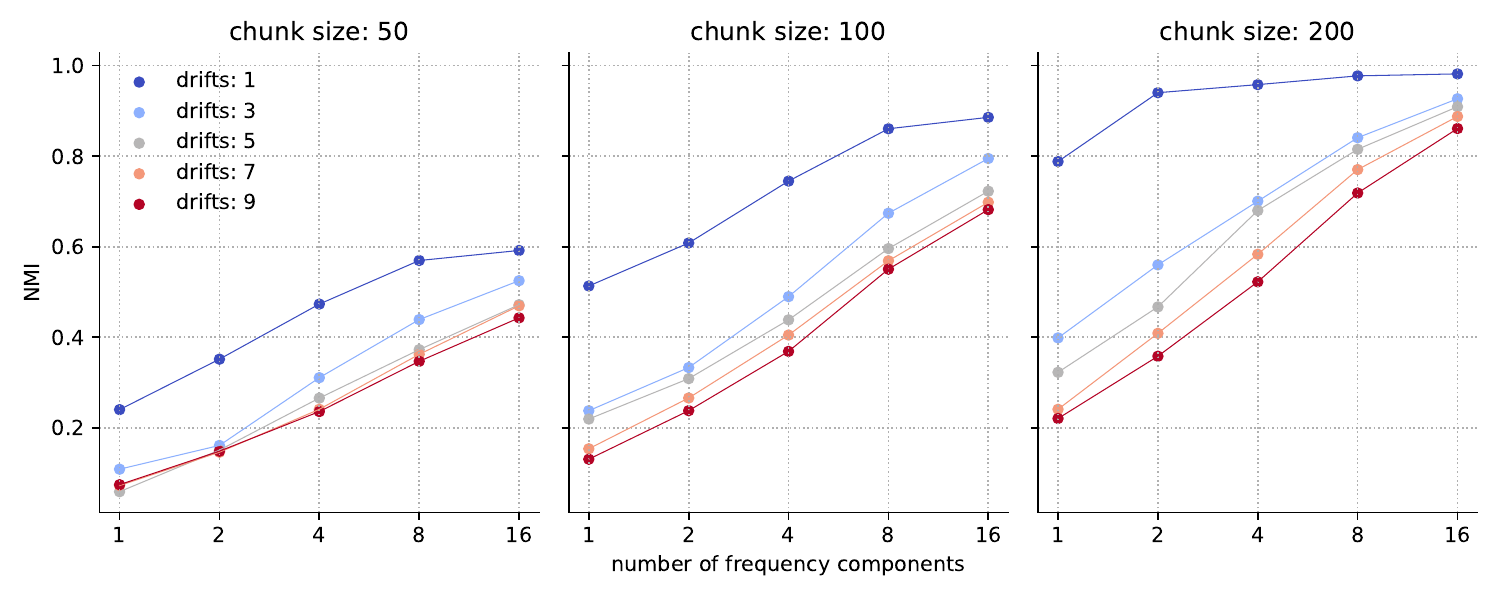}
    \caption{The relation between \textit{normalized mutual information} and~the~value of~$n$ hyperparameter for various chunk sizes (columns) and~various numbers of~drifts (line colors). The~values of~the~x-axis determine the~value of~the~$n$ hyperparameter.}
    \label{fig:e1}
\end{figure}

The first experiment evaluated the~values of~the~$n$ hyperparameter, describing the~number of~frequency components selected by the~\textsc{ffm} approach. The~averaged results are~presented in~Figure~\ref{fig:e1}, showing the~value of~\textit{normalized mutual information} between clusters describing the~concepts in~the data stream and~the~\textit{concept ground-truth}, describing the~actual concept. 

As expected, the~task of~concept clustering is~becoming more difficult with a~smaller number of~samples in~the~data chunk. It can result from less accurate frequency component selection or~the~generation of~less diverse frequency representation of~data. Therefore, when possible, selecting a~larger chunk size should allow for a~more precise data stream description. Similarly, the~task becomes more difficult when the~number of~concept drifts rises. The~growing number of~concepts while the~data stream length remains constant results in~fewer samples describing each cluster. This can result in~less accurate frequency selection or~increase the~complexity of~the~clustering task, where the~number of~formed clusters increases.

The results show that the~highest quality of~concept identification is~obtained for the~largest number of~frequency components $n=16$. However, for more straightforward scenarios (single concept drift and~large chunk size), the~clustering quality is~already high for $n=2$. Therefore, it can be expected that, in~some cases, the~value of~this hyperparameter can be reduced without a~significant drop in~recognition quality while limiting the~\textit{metadescription} dimensionality even further. 

\subsection{Comparison with reference approaches}
\begin{figure}[!b]
    \centering
    \includegraphics[width=0.8\linewidth]{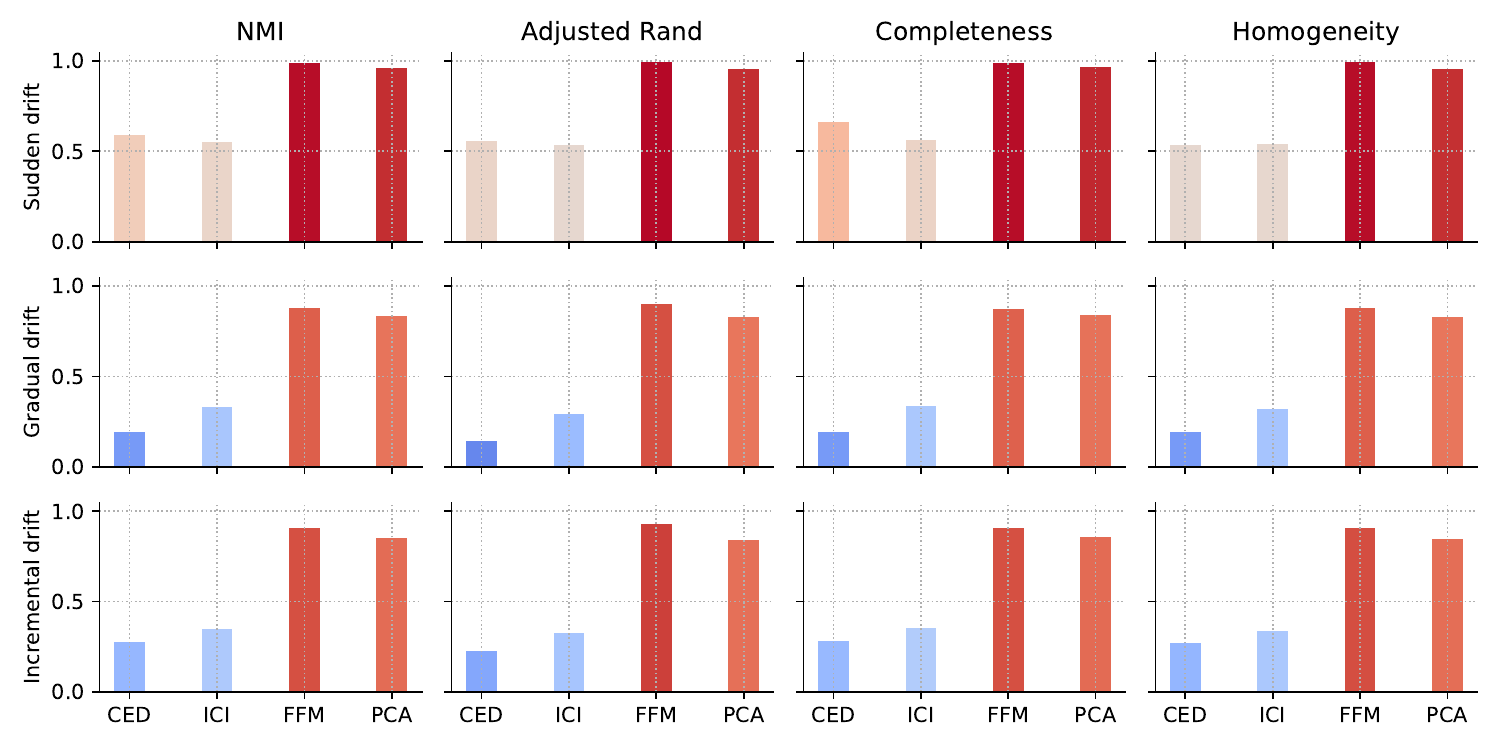}
    \caption{The results of~the~second experiments across four different metrics (in columns) and~for three considered types of~drifts (in rows). The~color of~the~bar plot is~dependent on~the~obtained metric value -- the~higher results are~closer to~red.}
    \label{fig:e2}
\end{figure}

The second experiment compared the~representation generated by \textsc{ffm} with two \textit{state-of-the-art} strategies and~the~baseline feature extraction with \textsc{pca}. The~results are~presented graphically in~Figure~\ref{fig:e2}.

The rows of~the~figure indicate the~dynamics of~concept drifts -- sudden, gradual, and~incremental, respectively. The~columns present the~concept identification quality with four metrics. The~height of~the~bar and~its color indicate the~average result of~a~particular representation. It is~visible that the~best results are~achieved by \textsc{ffm} and~\textsc{pca} methods, with a~slight advantage of~\textsc{ffm}. In~sudden concept changes, allowing for a~direct and~unequivocal separation of~concepts, the~quality of~\textsc{ced} and~\textsc{ici} metafeature sets is~similar. However, concerning gradual and~incremental concept changes, \textsc{ici} has a~slight advantage. Those continuous changes will result in~data chunks forming adjacent clusters, complicating the~separation process. In~a~complex setting, the~more extensive set of~features used in~\textsc{ici} results in~a~better concept separation.

\begin{table}[!htb]
    \centering
    \caption{The results of~the~\textit{metadescription} comparison. The~cells present averaged results, with their standard deviation in~parenthesis. The~indexes of~statistically significantly worse methods are~presented under each averaged result.}
    \setlength{\tabcolsep}{5pt} 
\renewcommand{\arraystretch}{1.1} 
\scriptsize

\begin{tabular}{l|l|cccc}
\toprule
                               &             & \textbf{CED}           & \textbf{ICI}            & {\color{red}\textbf{FFM}}               & \textbf{PCA}               \\
                               &             & (0)             & (1)              & {\color{red}(2)}                &  (3)                \\ \midrule
\multirow{6}{*}{\rotatebox[origin=c]{90}{\textsc{nmi}}}                         & \textsc{sudd}      & 0.595 (0.145) & 0.548 (0.203) & \textbf{0.991} (0.014)    & \textbf{0.960} (0.053)    \\
                               &             &   ---            &         ---      & 0, 1 & 0, 1 \\
                               & \textsc{grad}     & 0.190 (0.061) & 0.318 (0.127) & \textbf{0.876} (0.032)    & \textbf{0.834} (0.068)    \\
                               &             & ---           & 0             & 0, 1             & 0, 1             \\
                               & \textsc{incr} & 0.276 (0.115) & 0.347 (0.142) & \textbf{0.906} (0.023)    & \textbf{0.851} (0.078)    \\
                               &             & ---           & ---           & 0, 1             & 0, 1             \\ \midrule
\multirow{6}{*}{\rotatebox[origin=c]{90}{\textsc{adj. rand}}} & \textsc{sudd}      & 0.557 (0.164) & 0.531 (0.214) & \textbf{0.994} (0.011)    & \textbf{0.958} (0.082)    \\
                               &             &      ---         &        ---       & 0, 1 & 0, 1 \\
                               & \textsc{grad}     & 0.139 (0.045) & 0.273 (0.119) & \textbf{0.903} (0.032)    & \textbf{0.828} (0.116)    \\
                               &            &        ---       & 0       & 0, 1 & 0, 1 \\
                               & \textsc{incr} & 0.223 (0.100) & 0.324 (0.154) & \textbf{0.930} (0.024)    & \textbf{0.842} (0.126)    \\
                               &             &        ---       &            ---   & 0, 1 & 0, 1 \\\midrule
\multirow{6}{*}{\rotatebox[origin=c]{90}{\textsc{completeness}}}   & \textsc{sudd}      & 0.666 (0.165) & 0.558 (0.205) & \textbf{0.990} (0.014)    & \textbf{0.968} (0.035)    \\
                               &             &        ---       &            ---   & 0, 1 & 0, 1 \\
                               & \textsc{grad}     & 0.190 (0.060) & 0.326 (0.128) & \textbf{0.873} (0.033)    & \textbf{0.840} (0.059)    \\
                               &             &        ---       & 0       & 0, 1 & 0, 1 \\
                               & \textsc{incr} & 0.282 (0.122) & 0.356 (0.145) & \textbf{0.904} (0.024)    & \textbf{0.856} (0.068)    \\
                               &             &        ---       &            ---   & 0, 1 & 0, 1 \\ \midrule
\multirow{6}{*}{\rotatebox[origin=c]{90}{\textsc{homogeneity}}}  & \textsc{sudd}      & 0.538 (0.130) & 0.539 (0.203) & \textbf{0.991} (0.013)    & \textbf{0.954} (0.070)    \\
                               &             &        ---       &            ---   & 0, 1 & 0, 1 \\
                               & \textsc{grad}     & 0.190 (0.061) & 0.311 (0.128) & \textbf{0.878} (0.030)    & \textbf{0.830} (0.080)    \\
                               &             &          ---     & 0       & 0, 1 & 0, 1 \\
                               & \textsc{incr} & 0.271 (0.111) & 0.339 (0.140) & \textbf{0.908} (0.023)    & \textbf{0.847} (0.089)    \\
                               &             &     ---          &           ---    & 0, 1 & 0, 1 \\ \bottomrule 
\end{tabular}

    \label{tab:table}
\end{table}

Certain limitations of~performed experimental comparison are~worth mentioning here when focusing on~\textit{post-hoc} data stream analysis. Since \textsc{ced} and~\textsc{ici} are~capable of~incremental processing of~data streams, selecting specific metrics does not allow for preliminary evaluation of~their variance and~precise selection of~informative metafeature pool. Those \textit{state-of-the-art} metafeature combinations are~intended to~describe the~universal properties of~data. Meanwhile, \textsc{ffm} and~\textsc{pca} describe the~data with preliminary knowledge about the~variance in~the~concept distributions, which may substantiate their high outcomes. The~high results of~the~baseline \textsc{pca} reveal the~strengths of~feature extraction in~the~task of~data stream \textit{metadescription}. However, the~particular drawback of~the~\textsc{pca} approach is~the~lack of~metafeature semantics. While this feature extractor has already been used for drift detection task~\cite{agrahari2022adaptive}, the~extracted components do not hold a~precise interpretation. In~contrast, the~frequencies extracted with \textsc{ffm} precisely describe the~feature's correlations with periodic functions, allowing for their visualization and~even restoration of~an approximated feature vector.

Table~\ref{tab:table} presents the~quantitative results of~this experiment. The~columns represent the~evaluated methods, and~the~rows -- specific clustering metrics and~types of~concept dynamics in~the~data stream. The~table additionally presents the~results of~paired Student's t-test for independent samples with an $\alpha=5\%$. In~each row of~the~table, the~average result of~the~methods that are~statistically significantly better than the~largest number of~references are~emphasized in~bold. As~expected based on~results presented in~Figure~\ref{fig:e2}, the~\textsc{ffm} and~\textsc{pca} are~among the~best methods across all streams and~metrics. It is~worth noting that the~results of~\textsc{pca} have a~larger standard deviation, indicated in~parenthesis in~each cell.

\subsection{Identifying the~number of~concepts}

\begin{figure}[!b]
    \centering
    \includegraphics[width=0.9\linewidth]{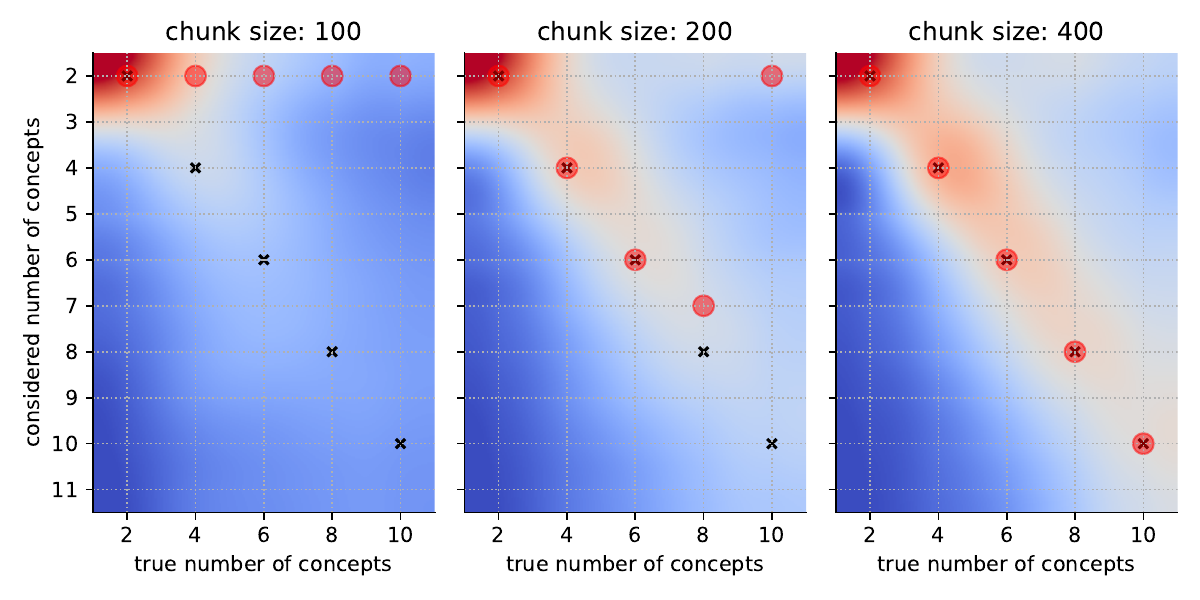}
    \caption{The results of~the~experiment in~the~form of~a~heatmap with interpolated values. The~background color describes the~average \textit{silhouette score} of~the~clustering task. The~horizontal axis shows the~true number of~concepts in~the~stream, and~the~vertical axis -- the~considered number of~concepts. The~red point identified the~number of~concepts with the~highest score for each processed stream type.}
    \label{fig:e3}
\end{figure}
 
The final experiment focused on~identifying the~number of~concepts in~the~stream and, additionally, graphically presented the~data chunks coming from various concepts. In~this experiment, only the~\textsc{ffm} approach was evaluated. The~graphical results are~presented in~Figure~\ref{fig:e3} for three evaluated chunk sizes, presented in~columns of~the~figure.

The heatmaps visible in~the~figure show the~\textit{silhouette score} of~the~clusters identified with \textit{k-means}. The~values close to~red indicate a~high metric value, indicating a~good clustering quality, and~close to~blue -- low \textit{silhouette score}. The~actual number of~concepts (i.e., the~stream type) is~visible on~the~horizontal axis, and~the~number of~concepts considered in~the~search is~on the~vertical axis. The~black markers indicate the~actual number of~clusters, and~the~red markers show the~result with the~highest score for each stream type. Ideally, the~red and~black markers should overlap -- indicating that the~actual number of~concepts was identified correctly based on~maximizing the~\textit{silhouette score}.

Such a~result is~visible for the~largest examined chunk size of~400 samples. The~chunk size of~100 did not allow for identifying the~number of~concepts since, regardless of~the~stream type, the~best score was obtained for two clusters. In~the~case of~a~chunk size equal to~200, the~number of~concepts was correctly identified for up to~six concepts present in~the~steam. As already noticed in~the~first experiment, the~larger chunk size results in~a~better \textit{metadescription} of~the~data stream.

\begin{figure}[!b]
    \centering
    \includegraphics[width=\linewidth]{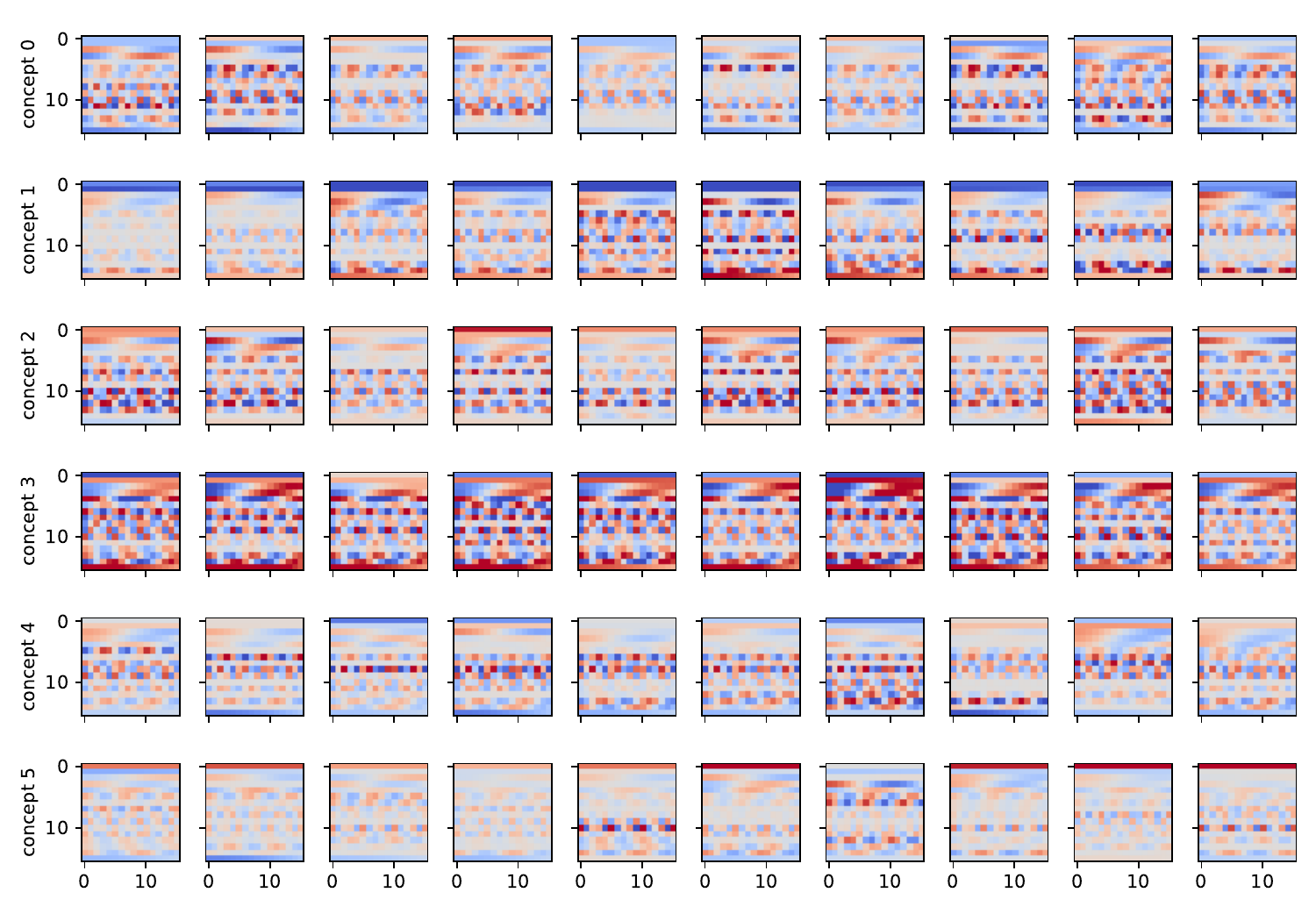}
    \caption{The visualization of~data chunks from six concepts present over the~course of~the~stream. Each row presents ten data chunks of~the~stream coming from various concepts.}
    \label{fig:e3_vis}
\end{figure}

To show the~complete set of~properties of~the~proposed method, the~visualization of~frequencies was presented in~Figure~\ref{fig:e3_vis}. The~presented data stream contained six concepts over 500 chunks of~size 400 and~a~dimensionality of~500. The~visualization of~chunks from particular concepts is~presented in~rows of~the~figure. Each image presented in~the~figure shows a~single data chunk -- with restored frequencies selected by \textsc{ffm} stacked in~rows. In~the~visualization, the~high component values are~presented in~red, and~the~low values in~blue. However, the~interpretation of~specific feature values can be shown with any pseudocolor mapping, enabling the~highlighting of~a~particular range of~values or~increasing their contrast. 

The visualization tool states a~valuable addition to~the~\textit{metadescription} generated with \textsc{ffm}, allowing for the~interpretation of~selected frequencies and~visual differentiation of~specific concepts.

\section{Real-world data stream concept identification}\label{sec-insects}

This section aims to~present the~utilities of~the~proposed approach in~the~\emph{explanation} of~changes visible in~the~real-world data streams. For this purpose, the~\textsc{insects} data streams~\cite{souza2020challenges} were analyzed in~a~framework following the~processing scheme of~(\textit{a}) extracting the~\textit{metadescription} of~a~data stream with \textsc{ffm} (\textit{b}) identifying of~the~number of~concepts present in~the~stream based on~\textit{k-means} clustering and~\textit{silhouette score} maximization, and~(\textit{c}) separating the~stream into specific concepts. The~description of~the~data streams is~presented in~Table~\ref{tab:insects}. Since the~\textsc{ffm} operates in~an unsupervised mode, the~class labels were discarded for the~time of~stream analysis.

\begin{table}[h]
    \caption{Characterization of~\textsc{insects} data streams}
    \centering
    \setlength{\tabcolsep}{6pt} 
    \renewcommand{\arraystretch}{1.1} 
    \scriptsize

    \begin{tabular}{l|r|r}
    \toprule
    \textsc{data stream name}                            & \textsc{number of~samples} & \textsc{chunk size} \\
    \midrule
    \textsc{insects} abrupt imbalanced                  & 355 000            & 500        \\
    \textsc{insects} abrupt balanced                    & 52 800             & 50         \\
    \textsc{insects} gradual imbalanced                 & 143 200            & 200        \\
    \textsc{insects} gradual balanced                   & 24 150             & 50         \\
    \textsc{insects} incremental imbalanced             & 452 000            & 500        \\
    \textsc{insects} incremental balanced               & 57 000             & 100        \\
    \textsc{insects} incremental recurring imbalanced & 452 000            & 500        \\
    \textsc{insects} incremental recurring balanced   & 79 900             & 100        \\
    \textsc{insects} incremental abrupt imbalanced      & 452 000            & 500       \\
    \textsc{insects} incremental abrupt balanced        & 79 900             & 100        \\
    \bottomrule

    \end{tabular}

    \label{tab:insects}
\end{table}

All used data streams are~described with 33 features. The~chunk size was selected depending on~the~number of~samples to~allow the~clear presentation of~the~entire steam, resulting in~a~selection of~a~larger chunk size for the~more significant number of~samples. Data stream processing in~a~batch mode resulted in~the~need to~discard the~samples that did not form an entire chunk. Hence, the~minor differences between the~original number of~samples of~the~data stream and~the~value presented in~the~table can be noticed. The~relatively small dimensionality of~the~problem resulted in~the~selection of~a~small number of~$n=5$ frequency components used for the~data stream \textit{metadescription}.

\begin{figure}[!t]
    \centering
    \includegraphics[width=0.49\linewidth]{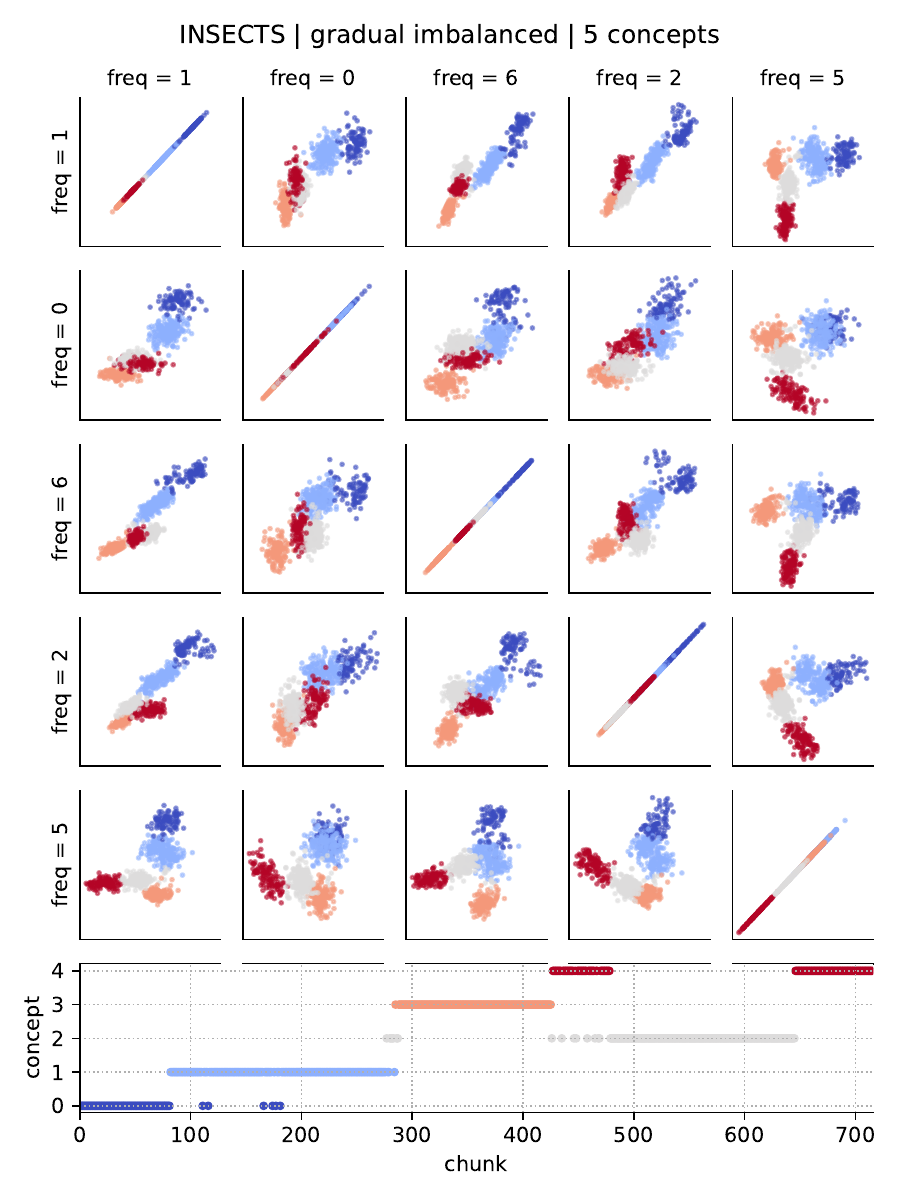}
    \includegraphics[width=0.49\linewidth]{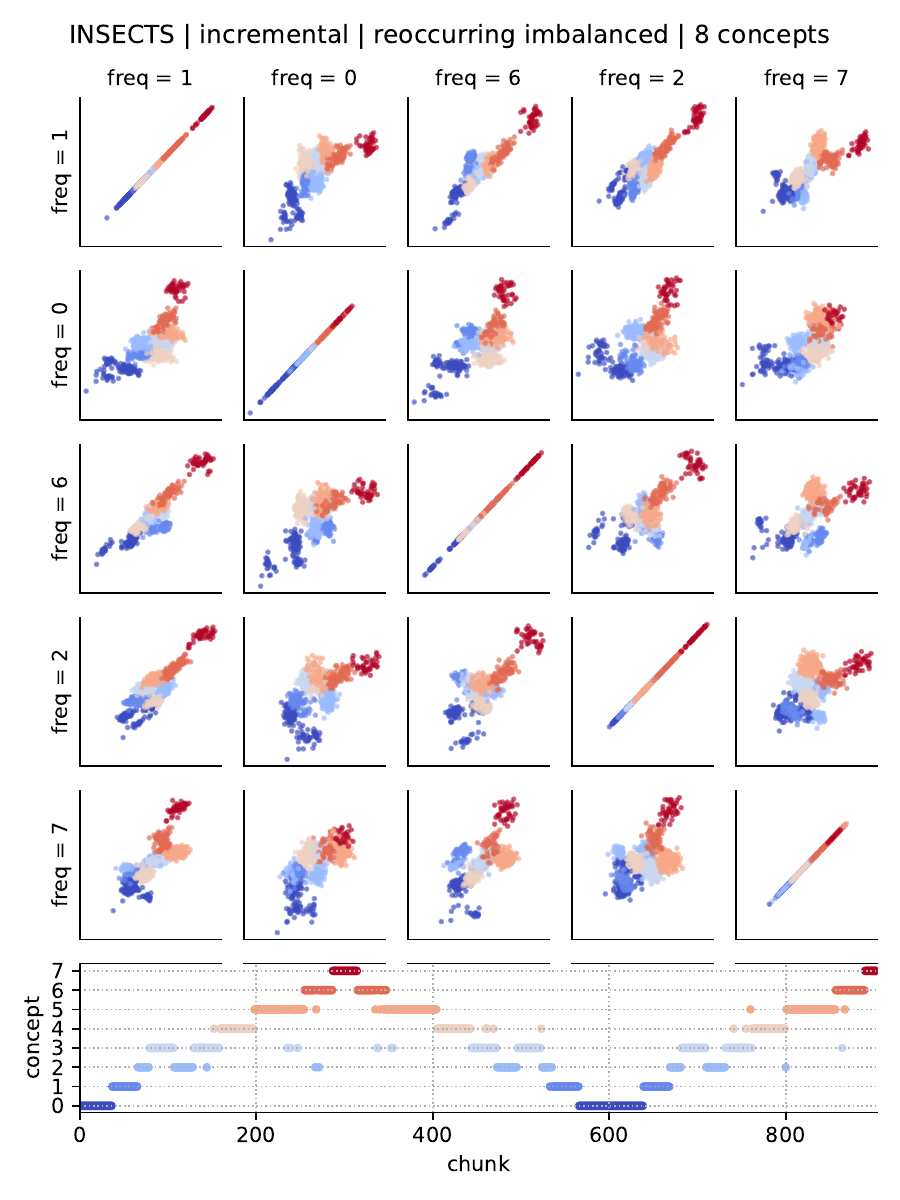}
    \caption{The representation of~selected \textsc{insects} data streams generated with \textsc{ffm} and~clustered into processed concepts (identified with colors). The~time of~concept occurrence is~shown in~the~bottom plot.}
    \label{fig:e4}
\end{figure}

Figure~\ref{fig:e4} presents the~results of~data stream \textit{metadescription} of~two selected \textsc{insects} data streams. The~scatter plots in~the~top part of~the~figure show the~location of~chunk representation in~the~multidimensional space describing the~frequency components of~the~data samples. The~colors indicate the~separation into specific concepts identified in~the~stream. 

The number of~concepts was indicated by \textit{silhouette score} maximization~\cite{rachwal2023determining}. The~presented processing scenario considered the~separation into from 4 to~10 concepts and~selected the~most promising value across the~evaluated ones. The~results of~clustering were stabilized with 10 replications of~the~clustering procedure. After the~separation, without considering the~time dependency of~samples, the~concept membership was presented at the~bottom of~a~figure -- where each chunk was assigned to~a~specific concept. The~figure allows to~notice the~smooth transition between concepts, especially in~the~\textit{incremental recurring} stream presented on~the~right side of~the~figure, where the~initial concept recurrence is~clearly visible around the~600th data chunk. Not considering the~time dependency of~samples indicates the~lack of~direct assumption that the~consecutive data chunks represent similar concept. Without such an assumption, the~fact that most of~the~concepts span across the~adjacent data batches can substantiate that the~\textsc{ffm} provides a~high-quality data stream \textit{metadescription}. 

The complete results of~concept separation are~presented in~Table~\ref{tab:insects-res} in~three metrics: \textit{silhouette} (\textsc{sil}), \textit{calinski-harabasz} (\textsc{c-h}) and~\textit{davies-bouldin} (\textsc{d-b}) scores. Those metrics measure the~internal quality of~concept separation without the~requirement of~\textit{concept drift ground-truth}, not available for most of~the~the real-world data streams~\cite{bifet2015efficient}.

\begin{table}[]
    \caption{Averaged results of~concept separation in~the~\textsc{insects} data streams and~the~number of~identified concepts}
    \centering
    \setlength{\tabcolsep}{5pt} 
    \renewcommand{\arraystretch}{1.1} 
    \scriptsize

    \begin{tabular}{l|r|r|r|r}
    \toprule
     \textsc{data stream}   & \textsc{concepts}  & \textsc{sil score} & \textsc{c-h score}  & \textsc{d-b score} \\
        &  & \textit{(maximize)} & \textit{(maximize)}  & \textit{(minimize)} \\
    \midrule
    \textsc{insects} abrupt imbalanced                  & 5 & 0.458  & 794.747  & 0.863  \\
    \textsc{insects} abrupt balanced                    & 5 & 0.308  & 611.757   & 1.025  \\
    \textsc{insects} gradual imbalanced                  & 5 & 0.531  & 1347.013  & 0.680  \\
    \textsc{insects} gradual balanced                    & 4 & 0.438  & 435.022   & 0.901  \\
    \textsc{insects} incremental imbalanced             & 8 & 0.431  & 834.275  & 0.812 \\
    \textsc{insects} incremental balanced               & 5 & 0.284  & 374.808  & 1.141  \\
    \textsc{insects} incremental recurring imbalanced & 8 & 0.432  & 868.377  & 0.813  \\
    \textsc{insects} incremental recurring balanced   & 7 & 0.382  & 836.204  & 0.930 \\
    \textsc{insects} incremental abrupt imbalanced    & 8 & 0.421  & 837.635  & 0.829  \\
    \textsc{insects} incremental abrupt balanced      & 5 & 0.424  & 1341.345  & 0.797  \\
    \bottomrule
    \end{tabular}

    \label{tab:insects-res}
\end{table}

The better concept clustering is~indicated by a~higher \textit{silhouette} and~\textit{calinski-harabasz} scores, and~the~lower \textit{davies-bouldin score}. The~presented results intend to~allow for comparison with future reference approaches. Those metrics aim to~\textit{estimate} the~internal quality of~concept separation and~are characterized with an inevitable bias~\cite{rachwal2023determining,gao2018understanding}. It is~worth keeping in~mind that the~analysis of~concept membership in~real-world data streams, especially with non-sudden concept changes, is~always characterized by some uncertainty. The~direct separation of~smooth concept transition into discrete concepts highlights the~limitations of~the~real-world data stream evaluation.

\section{Conclusions}\label{sec-conclusions}

This work proposes a~tool for analyzing the~high dimensional data streams in~the~frequency domain, allowing for \textit{post-hoc} concept identification and~visualization of~the~data stream. The~proposed \textit{Frequency Filtering Metadescriptor} (\textsc{ffm}) searches for frequency components with the~largest variance across the~processed data chunks, allowing for (\textit{a}) generating the~frequency representation of~data with significantly reduced dimensionality, (\textit{b}) clustering of~processed data chunks into groups describing specific concepts and~(\textit{c}) visualization of~frequencies visible in~the~processed data chunks. 

The~presented experiments showed that the~proposed approach allows for concept identification competitive with the~baseline of~\textsc{pca} feature extraction and statistically significantly better than \textit{state-of-the-art} adapted in the concept drift detection and~data stream classification tasks. The~particular benefit of~the~proposed \textsc{ffm} over \textsc{pca} is~the~semantic meaning of~the~extracted metafeatures. In~the~final experiment, the~\textsc{ffm} method showed the~ability to~identify the~number of~concepts present in~the~data stream. This strategy was used to~analyze the~real-world \textsc{insects} data streams, showing promising results of~concept separation.

In future research, the~selected frequencies can be used to~reconstruct the~original feature vector, where the~dimensionality of~the~data needs to~be significantly reduced while the~original feature interpretation is~required. Adapting the~frequency analysis in~the~incremental learning scenario states an interesting future direction since, intuitively, the~processing in~the~frequency domain may offer additional generalization possibilities, critical for many machine learning tasks.

\subsection*{Acknowledgments}

This work was supported by the~statutory funds of~the~Department of~Systems and~Computer Networks, Faculty of~Information and~Communication Technology, Wrocław University of~Science and~Technology, as well as partially funded by the~National Center for Research and~Development within INFOSTRATEG program under the~number INFOSTRATEG-I/0019/2021-00.

\bibliographystyle{unsrt}
\bibliography{biblio}

\end{document}